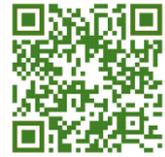

**Research Article**  **Open Access (CC–BY-SA)**

# Identification of chicken egg fertility using SVM classifier based on first-order statistical feature extraction

**Shoffan Saifullah [a,1,\*]; Andiko Putro Suryotomo [a,2]**

[a] Department of Informatics, Universitas Pembangunan Nasional Veteran Yogyakarta, Jl. Babarsari 2 Yogyakarta, 55283, Indonesia
[1] shoffans@upnyk.ac.id; [2] andiko.ps@upnyk.ac.id
\* Corresponding author

**Abstract**

This study aims to identify chicken eggs fertility using the support vector machine (SVM) classifier method. The classification basis used the first-order statistical (FOS) parameters as feature extraction in the identification process. This research was developed based on the process's identification process, which is still manual (conventional). Although currently there are many technologies in the identification process, they still need development. Thus, this research is one of the developments in the field of image processing technology. The sample data uses datasets from previous studies with a total of 100 egg images. The egg object in the image is a single object. From these data, the classification of each fertile and infertile egg is 50 image data. Chicken egg image data became input in image processing, with the initial process is segmentation. This initial segmentation aims to get the cropped image according to the object. The cropped image is repaired using image preprocessing with grayscaling and image enhancement methods. This method (image enhancement) used two combination methods: contrast limited adaptive histogram equalization (CLAHE) and histogram equalization (HE). The improved image becomes the input for feature extraction using the FOS method. The FOS uses five parameters, namely mean, entropy, variance, skewness, and kurtosis. The five parameters entered into the SVM classifier method to identify the fertility of chicken eggs. The results of these experiments, the method proposed in the identification process has a success percentage of 84.57%. Thus, the implementation of this method can be used as a reference for future research improvements. In addition, it may be possible to use a second-order feature extraction method to improve its accuracy and improve supervised learning for classification.

## Introduction

The livestock and poultry industry is still increasing productivity by using technology. The increasingly sophisticated technology makes researchers develop a lot, especially in identifying chicken egg fertility [1], [2]. The reference's relevance in this study is identification using image processing to the identification. Identification of egg fertility is carried out while the eggs are in the incubator [3]. This process runs during the first week of new eggs at hatching [4], separating fertile and infertile eggs [5]. If the seventh day of detection [6] shows infertile eggs, the eggs can be taken and consumed. Thus, in this identification process, in addition to detecting the fertility of chicken eggs at the beginning of the hatching process, infertile eggs can be used for other valuable purposes. This identification is mainly processed manually, although there are also uses of technology. However, the application of technology in identifying egg fertility (especially in Indonesia) is still being developed [7]. Thus, it is found in several articles that discuss the identification of chicken egg fertility or early embryo detection.

Many studies use image processing in identifying the fertility of chicken eggs [8]. The research developed various concepts such as image acquisition [9], image preprocessing [10], [11], and segmentation [11], [12]. Image acquisition development uses several tools, such as thermal camera [13], digital camera [14], hyperspectral [15], Arduino (based camera) [16]. Based on these tools, this study uses a dataset of images captured from a smartphone camera (a concept like a digital camera) [9], [17]. Thus, the result of the object is a replication of the real object. The dataset acquisition process uses irradiation on a single egg object (candling). Image preprocessing improves image quality to get better input. One of the methods is histogram equalization (HE), with its development using contrast limited histogram equalization. In addition, the feature extraction method uses second-order statistics. The identification process uses supervised learning such as Artificial Neural Network [8] and unsupervised learning such as K-means clustering [18].

The identification processes that have been developed have many excellent system accuracies. However, the development of this concept is still being researched to get maximum results and can be implemented in the detection of fertility in hatching eggs. This study focuses on identifying the fertility of chicken eggs with datasets from Saifullah et al. In addition, this study uses the FOS feature extraction method. FOS can provide the extracted features in five categories: mean, entropy, variance, skewness, and kurtosis. The calculation results of all these parameters are trained and tested using the SVM classifier method. This method identifies the image of chicken eggs into two classes, namely fertile and infertile.

This article consists of four main sections. This section (first) explains the background of the problem, related to reviews of previous research, gab, and the purpose of this research. Then, in the second part, we explain the methods used in this study, including the proposed method. The third part discusses the experimental results and discusses the things obtained in this study. Finally, the fifth section explains the conclusions of the experiments that the researchers have carried out.

**Method**

This study uses three main methods: image processing, feature extraction, and identification using the SVM classifier. These methods are used to identify the fertility of chicken eggs. The basic concept of image processing is to get the best image in obtaining feature extraction. Furthermore, feature extraction is a feature of an image or specific information used as a reference for identification. In this study, feature extraction used the first-order statistical (FOS) method. The parameters in the FOS become input for the identification process using the support vector machine (SVM) classifier. A detailed image of the identification process is shown in Figure 1.

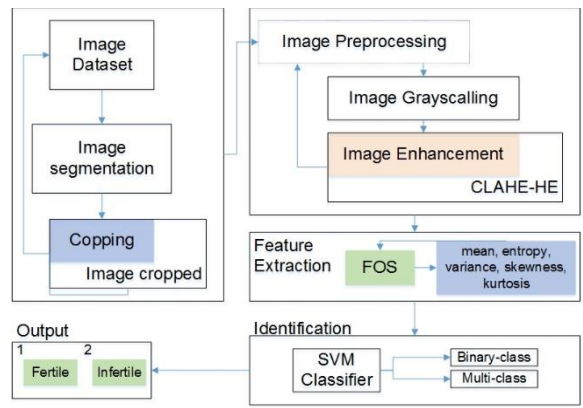

**Figure 1.** The steps of the identification process of Chicken's egg fertility based on FOS and SVM classifier

Figure 1 shows the steps of the identification process for the fertility of chicken eggs, which consists of five blocks, namely initial dataset processing, image preprocessing, feature extraction, identification, and the resulting output. This section generally describes these processes regarding the concepts and formulas used or the algorithms.

**A.  *Image processing on egg fertility identification***

Image processing is a method that processes images as input and is processed by various methods. Image processing: several of them are image acquisition, image preprocessing, and image segmentation [19]. Image acquisition is the first step to get a digital image. This study uses the acquisition process according to [8]. This process used a smartphone camera whose object is illuminated by LED lights (candling), as shown in Figure 2. (a). The data sample used in this study is the image of chicken eggs, totaling 100 data consisting of 50 fertile and 50 infertile eggs. Each image has a single egg object with a dark/black background (Figure 2. (b)).

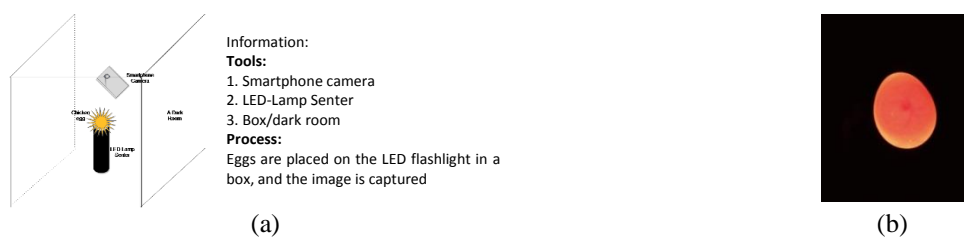

(a)  (b)

**Figure 2.** The process of image acquisition of a chicken egg is by (a) the system design that produced (b) an RGB image of the egg

In this study, based on Figure 1, the image (dataset) is processed using the primary segmentation method to detect images and perform cropping. So that the image results obtained from the dataset are images with only object images.

This process removes the background, which hampers both processing time and file size when processed directly. In detail, the cropping process uses area detection where data that is not worth 0 or close to will be marked and after all new processes are marked and cropped [11]. This concept uses the segmentation method, making it easier to represent images for analysis [20]. In this study, image segmentation divides the image into two parts: the object and the background. Segmentation is carried out by thresholding, edge detection, region-based, and others [21]. The segmentation used is the thresholding method which divides the color area [22] based on the difference in brightness (dark/light). The dark or near image area is converted to a black image with a value of 0.

The next stage is image preprocessing that used two methods, namely grayscaling and image enhancement. Grayscaling is a method used to convert color images into grayscale images. In image processing, color images have three color components, namely red (R), green (G), and blue (B) [23]. Convert color (RGB) images to grayscale images using the average formula of the three components or using (1) [24]. This study uses (1) with each color component having its multiplier weight.

$$G' = 0{,}2989 * R + 0{,}587 * G + 0{,}1141 * B \qquad (1)$$

After the grayscaling process, the image is enhanced [25] by the CLAHE-HE method. Combining the two methods gives an image improvement result that is close to optimal when compared to using one [10]–[12]. CLAHE is an improvement from HE using a limit value (the maximum height of the histogram generated by the image). HE is an image enhancement that uses histogram alignment of all pixels (n). The value of each gray pixel i is divided by the total gray pixel, where 0 represents black and 1 represents white. Meanwhile, CLAHE uses the calculation of the maximum limit value of the histogram height, which is calculated using (3), also called the clip limit calculation. In the CLAHE calculation, the calculated factors include area size (M), grayscale value (N), and clip factor (a), with a value range between 1 to 100.

$$hi = \frac{n_i}{n}, i = 0, 1, 2, \dots, L - 1 \qquad (2)$$

$$\beta = \frac{M}{n}\left(1 + \frac{\alpha}{100}(s_{max} - 1)\right) \qquad (3)$$

$$hybrid\ CLAHE - HE = \beta \oplus h_i \qquad (4)$$

The final result of image preprocessing is a grayscale image that has been improved with various calculations. The grayscale image has a single histogram which is used in calculating the value of the FOS parameters.

### B. First-order statistical feature extraction based on image

Feature extraction is a process to obtain information on particular image objects. Each image will have different information from other images. This extraction produces features as input parameters to distinguish specific objects in the identification or classification process. The feature extraction uses the FOS method. This method extracts features [26] using an image histogram to perform calculations [27]. FOS memiliki lima parameter untuk mendapatkan ciri suatu objek. Adapun kelima parameter tersebut adalah [28] Mean, Entropy, Variance, Skewness, and Kurtosis [29], [30]. Each parameter will be calculated based on the grayscale image from the previous preprocessing stage. In detail, the calculation process for each parameter (5)-(9).

Parameter calculation used a histogram from the grayscale image. Mean ($\mu$) uses a calculation method based on the size of the dispersion of an image. This calculation uses the value of the histogram and its grayscale matrix (5). Meanwhile, entropy (H) calculates to determine the irregularity of the shape of an image, which process uses the histogram value and its log calculation (6).

$$\mu = \sum_{n=0}^{N} f_n p(f_n) \qquad (5)$$

$$H = -\sum_{n=0}^{N} P(f_n)\ ^2 \log P(f_n) \qquad (6)$$

In addition to these two parameters, three other parameters use calculations based on histogram values and their continuation. The variance ($\sigma$) is the variation of each element in the image histogram and is also known as the standard deviation. Variance applies a calculation based on the grayscale value, histogram, and mean to calculate the process (7).

$$\sigma^2 = \sum_{n=0}^{N}(f_n - \mu)^2 P(f_n) \qquad (7)$$

Skewness ($a_3$) calculates the parameters using the components of the variance and the variance cube or the derivative of the standard deviation. This parameter is used to calculate the relative slope or slope of the image histogram curve (8). At the same time, the last parameter is kurtosis ($a_4$). This parameter is used to calculate the relative sharpness of the image histogram curve based on the 4th power of the standard deviation and standard deviation divisor (9).

$$a_3 = \frac{1}{\sigma^3} \sum_{n=0}^{N} (f_n - \mu)^3 P(f_n) \tag{8}$$

$$a_4 = \frac{1}{\sigma^4} \sum_{n=0}^{N} (f_n - \mu)^4 P(f_n) - 3 \tag{9}$$

*C. SVM classifier for identification*

SVM classifier is an algorithm used in pattern recognition and classification. This algorithm uses the concept of machine learning (supervised learning) with the principle of structural risk minimization (SRM). However, SVM has the basic principle of the linear classifier with linear and non-linear cases (in its development). SVM can be implemented to find the best hyperplane in classifying two classes. The pattern formed by the SVM is a member of the two classes +1 and -1 and shares alternative discrimination boundaries. Margin is the distance that separates the hyperplane from the closest pattern of the class. The closest pattern is the support vector, so the search for the best hyperplane is used as the core of SVM training. SVM is based on theory based on statistical learning with accuracy depending on certain kernel functions and parameters. Traditional SVM has a convention of class label values are +1 and -1 [31].

SVM classifies data linearly can be done by transforming the data into a high-dimensional feature space. These features affect significant computing processes. However, SVM introduces a kernel trick method based on Mercer's theorem (10). Equation (10) has limit values $a_j \geq 0$ and $\sum_j a_j y_j$ jajjyj. Besides, this equation is quadratic programming in optimization problems. Vector a can affect the value of w, so it needs to be processed by calculating the dot product of point pairs (11).

$$arg \max_a \sum_j a_j - \frac{1}{2} \sum_{j,k} a_j, a_k y_j y_k (x_j . x_k) \tag{10}$$

$$h(x) = sign\left(\sum_j a_j y_j (x . x_j - b)\right) \tag{11}$$

Based on these calculations, the weight j associated with each point is zero except for the supporting vector with the nearest dividing point. SVM training uses a dataset that has been labeled with a target class (e.g., two targets) from both classes and is classified as new data [32]. The limiting process can apply formula (12). The values of w and m are obtained by solving the optimization problem with (13). After this optimization, the parameter c is the penalty parameter, and ε is the error value using $lab_j(w.x_i) \geq 1 - \varepsilon_j, j = 1, 2, 3, \ldots, n$

$$y(x) = w.x + m \tag{12}$$

$$minimize \left(\frac{1}{2}|w|^2 + c \sum \varepsilon_j\right) \tag{13}$$

*D. Accuracy calculation using the confusion matrix*

Calculation of accuracy using the concept of accuracy. Accuracy is a process that determines data validation based on calculations between the results of measuring the actual value (closeness) [33]. This calculation used the confusion matrix with (14).

$$accuracy = \frac{TP+TN}{TP+TN+FP+FN} \tag{14}$$

True Positive (TP) is positive data that has a true predictive value. True Negative (TN) is negative data that has a correct prediction. False Positive (FP) is a Type I Error, where the data is predicted to be negative but has an optimistic prediction. Meanwhile, False Negative (FN) is a Type II Error, where the data is positive but has a negative prediction.

**Results and Discussion**

This section presents the results and discussion of the researcher's experiments. The results presented improve image quality with image processing, calculate samples from FOS, and identify results using the SVM classifier. In addition, we also complete the test and get accurate results.

*A. Image processing results for feature extraction input*

In this study, the initial process is image acquisition which produces a color image (RGB) with a single egg object (Figure 2. (b). The image is segmented using the primary method to distinguish the image from the object. So the results obtained are as in Figure 3. Figure 3. (a) is the result of image segmentation analysis using the concept of a binary image or black and white image. This initial process converts the image first using the binary method with the intensity value < 125 being changed to 0 or black and the other (> 125) being converted to 1. It is then reversing the value to indicate the presence of the object that has been obtained. Thus, based on Figure 3. (a), the middle part is the egg image object that is visible. After that, the image is cropped based on the defined boundary area around the object.

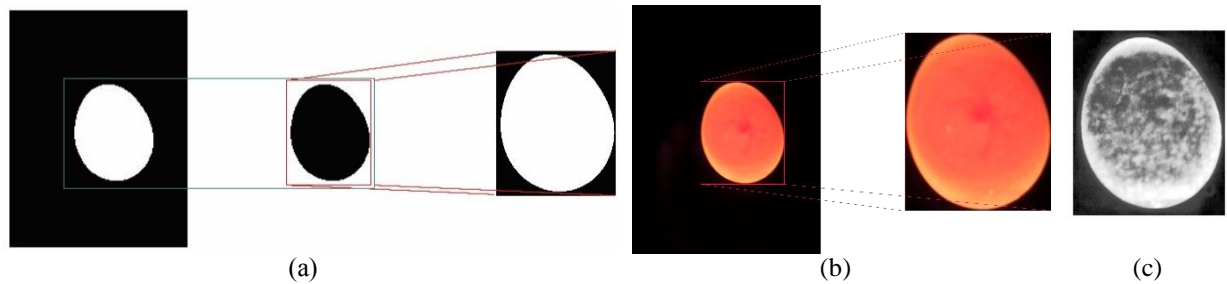

**Figure 3.** Image processing results to produce cropped images using segmentation methods (a) object detection, (b) color image cropping, and (c) image conversion to grayscale (grayscaling).

Proses copping tersebut, awalnya dicobakan pada citra hitam putih. Kemudian, citra asli (hasil akusisi) dicrop seperti pada citra hitam putih tersebut. Sehingga hasil citra yang digunakan adalah citra tercrop deperti pada Gambar 3. (b) bagian kanan. Selain mempercepat proses, proses ini digunakan untuk mendapatkan inforamsi detail mengenai ciri objek. Sehingga dalam proses kalkulasinya mampu memberikan nilai terbaik untuk proses klasifikasi. Sebelum diekstrasi ciri, Citra tersebut diperbaiki pada image preprocessing menggunakan metode image enhancement.

The copping process was initially tested on black and white images. Then, the original image (acquisition result) is cropped as in the black and white image. So that the result of the image used is the cropped image as shown in Figure 3. (b) the right side. In addition to speeding up the process, this process is used to get detailed information about the characteristics of objects. The calculation process can provide the best value for the classification process. Before feature extraction, the image is corrected in image preprocessing using the image enhancement method.

Image preprocessing in this research uses only two methods: grayscaling, which is processed using (1). Moreover, the following process is an image enhancement using the CLAHE-HE method. The results of image preprocessing are shown in Figure 3. (c). This image is the last image used in calculating FOS parameters.

### B. First-order statistical feature extraction result and the calculation

Based on the results of image preprocessing, grayscale images are used as input for feature extraction calculations. The grayscale image has a histogram with a single color value between 0 and 255. Using the image enhancement method, the improved grayscale image has a histogram following the limits calculated by this method. Thus, implementing this FOS calculation produces values influenced by the histogram (as in Figure 4).

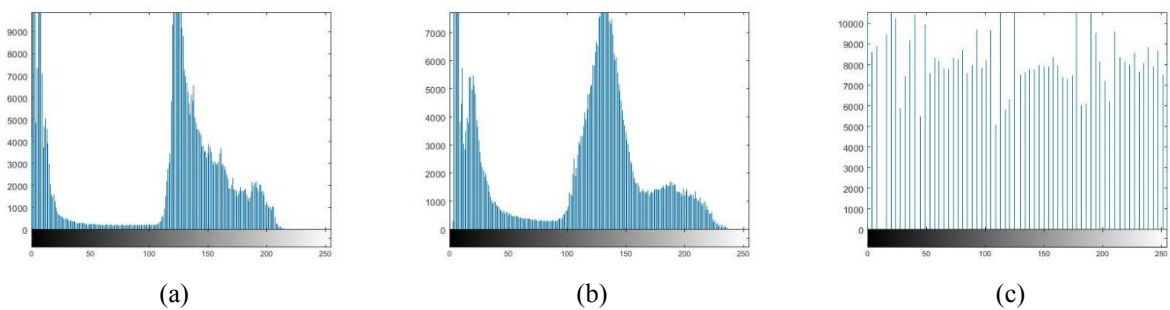

**Figure 4.** Image histogram generated from Figure 3. with (a) original histogram (grayscale), (b) CLAHE corrected histogram, (c) CLAHE-HE corrected histogram

Based on the image processing, Figure 4 shows the histogram results obtained from the original image (Figure 4. (a)). The image has the original distribution value of the input image. As for Figure 3. (b) is an image histogram processed using the CLAHE method, so the histogram value is lower than the original. This histogram shows that the improved image with CLAHE reduces image values and makes the image more contrasting. Furthermore, Figure 4. (c) is a histogram image based on the HE process from the CLAHE image. The process spreads the value of the image as a whole. The range of values is 0 to 255, with the histogram height close to the same. Based on the three histograms, we use Figure 4. (c) following previous studies, which showed that the improved image with the CLAHE-HE method was better than the single method [10]–[12].

Based on the distribution of the histogram values in Figure 3. (c), the calculation of the FOS parameter uses (5)-(9). The results of the calculation sample from FOS for the dataset use a sample of 10 data presented in Table 1. The data presented uses 5 data from fertile and infertile eggs, respectively. From the ten samples, it shows that the two parameters have different value distributions. However, some calculated parameters have nearly the same value and variation (larger or less). Values with different variations are mean, entropy and variance.

In contrast, the other two parameters have indeterminate values. However, the entropy value can also have different variations, in some cases being tested singly. However, we still use these parameters for parameter input, considering that although the values vary, we see that there are several closes and clustered values when grouped. So this is used as a reference to keep considering and using these parameters.

**Table 1.** The result of calculating FOS parameter values with initial identification

| Data | Mean | Entropy | Variance | Skewness | Kurtosis | Detected |
|---|---|---|---|---|---|---|
| 1 | 102.03 | 6.72 | 4332.69 | -0.50 | -1.23 | Fertile |
| 2 | 103.89 | 6.70 | 4447.91 | -0.57 | -1.27 | Fertile |
| 3 | 103.70 | 6.45 | 4155.24 | -0.69 | -1.23 | Fertile |
| 4 | 100.65 | 6.35 | 4291.52 | -0.61 | -1.35 | Fertile |
| 5 | 104.95 | 6.55 | 4370.24 | -0.64 | -1.23 | Fertile |
| 6 | 41.38 | 5.98 | 719.20 | -0.24 | -1.19 | Infertile |
| 7 | 20.28 | 5.54 | 222.40 | 0.51 | -0.56 | Infertile |
| 8 | 25.14 | 5.67 | 208.06 | 0.12 | -1.01 | Infertile |
| 9 | 20.56 | 5.49 | 237.13 | 0.48 | -0.81 | Infertile |
| 10 | 27.10 | 4.89 | 228.89 | -0.19 | 0.51 | Infertile |

The calculation results for each parameter for the 10 data indicate that each of the five fertile and infertile data can have a pattern, as shown in Figure 5. The eggs identified as fertile have a pattern with a high value, as shown in the picture with the fertile description. From the value of the pattern, the average change looks like a parallelogram pattern, both for fertile and infertile eggs. However, what makes the difference is the resulting value. Parameters of infertile eggs have a low-value calculation.

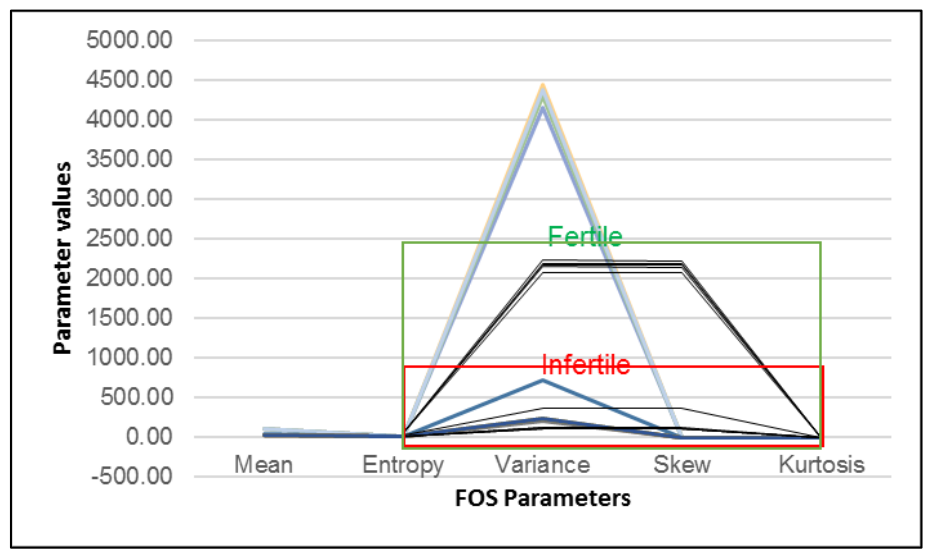

**Figure 5.** The pattern of parameter values generated by using data from actual calculations from both images of eggs detected as fertile and infertile

The FOS result is the result with the actual value of the calculation. In this case, when implementing the data normalization process, the data will be converted into more minor data with a constant value between 0 and 1. In this study, the classification results are less than optimal when implemented with normalization because the values are uniform. In addition, classification with actual data has a higher level of accuracy than normalized data. That is the reason why this research uses actual data.

## C.   Result of SVM classifier for chicken egg fertility identification and the accuracy

The results of testing using these methods are discussed in this section. In this study, identifying the fertility of chicken eggs classified eggs into two categories, namely fertile and infertile. The dataset used is 100 images of a single chicken egg image. The distribution of test and training data is 50 data each. Scenario testing is carried out using training samples, with the test results being 100% able to recognize fertile and infertile eggs. The following testing process is to use test sample data in addition to training data. The testing process is carried out sequentially and varies with scenarios such as Table 2.

Based on the experiments conducted, the researchers calculated the accuracy of the system's success using (14). The calculation process uses five accuracy-checking scenarios. The details of the five scenarios are shown in Table 2. The testing process uses a sample of 50 data, with the test scenario adding 10 data starting with the initial 10 data. These results were conducted to determine the percentage of testing both small data and changes. Based on the data that has been tested, the best level of image accuracy is the fertile egg data, in which the accuracy value of the test results is greater than the infertile egg data.

Table 2. Test Results and accuracy calculations with five scenarios

| Number of Data | Fertile | Infertile | Detection | Accuracy (%) |
|---|---|---|---|---|
| 10 | 5 | 4 | 9 | 90.00% |
| 20 | 9 | 6 | 17 | 85.00% |
| 30 | 14 | 9 | 25 | 83.33% |
| 40 | 19 | 12 | 33 | 82.50% |
| 50 | 24 | 15 | 41 | 82.00% |
| Average of accuracy | | | | 84.57% |

After analyzing the testing process, several factors affect the results of this accuracy test. These factors include infertile eggs with characteristics close to the same as fertile, which are classified as fertile or vice versa. From these scenarios, the initial data shows that the failure to detect infertility is 10%, while the detection of fertile is 0%. Along with new test data, the average percentage of fertile tests has a failure of 5%. Meanwhile, the system failed to identify infertility has an average accuracy of 36%. However, overall the system can identify the fertility of chicken eggs with an accuracy of 84.57% success. Each test's accuracy is above 82%.

In addition, another factor is the poor quality of the acquisition results. It is shown during image processing which has limitations in the visible presence or absence of embryos after preprocessing. In addition, the possibility of image preprocessing used is still not optimal. Alternatively, it can also be caused by unsuitable feature extraction, so selecting the features used in identification is necessary.

**Conclusion**

This study presents the FOS and SVM classifier methods to identify the fertility of chicken eggs. In addition, this process uses image processing techniques where the image results are used as input in extracting image features. The sample used in the previous research by Saifullah et al., both from the acquisition process and the image preprocessing process. Next is the improvement of the method using FOS feature extraction and SVM classification. The application of these two methods can identify the fertility of chicken eggs. The identification process using five input parameters has an average accuracy of 84.57%. This test can be used to reference that the system has a weakness in using the features obtained from the image. In addition, another possibility is that several datasets used between fertile and infertile eggs have the same extraction results, so that they are challenging to identify or use simple image processing methods so that the resulting image is not optimal. However, the accuracy results can be used as a reference for improvements in image processing, feature extraction, and classification. Image processing improvements aim to improve the quality of the acquired and preprocessed images. The feature extraction can be improved by using second-order statistics or textures. Meanwhile, in terms of classification, the development of other supervised learning methods can be used.

**Acknowledgment**

We thank Universitas Pembangunan Nasional Veteran Yogyakarta (especially the Department of Informatics) and LPPM UPN Veteran Yogyakarta to publish our research. Moreover, they have provided funding for this article. Funding for this research comes from the grand research "PDP" with agreement number B/41/UN.62/PT/IV/2021.